\documentclass[letterpaper]{article} % DO NOT CHANGE THIS
\usepackage{aaai24}  % DO NOT CHANGE THIS
\usepackage{times}  % DO NOT CHANGE THIS
\usepackage{helvet}  % DO NOT CHANGE THIS
\usepackage{courier}  % DO NOT CHANGE THIS
\usepackage[hyphens]{url}  % DO NOT CHANGE THIS
\usepackage{graphicx} % DO NOT CHANGE THIS
\usepackage{natbib}  % DO NOT CHANGE THIS AND DO NOT ADD ANY OPTIONS TO IT
\usepackage{caption} % DO NOT CHANGE THIS AND DO NOT ADD ANY OPTIONS TO IT
\usepackage{algorithm}
\usepackage{subfig}
\usepackage{bm}
\usepackage{tabularx}           
\usepackage[noend]{algpseudocode}
\usepackage{amssymb}
\usepackage{dsfont}
\usepackage{amsmath}
\newcolumntype{L}[1]{>{\raggedright\arraybackslash}m{#1}}
\newcolumntype{P}[1]{>{\centering\arraybackslash}p{#1}}
\usepackage{newfloat}
\usepackage{listings}
\DeclareCaptionStyle{ruled}{labelfont=normalfont,labelsep=colon,strut=off} % DO NOT CHANGE THIS
\floatstyle{ruled}
\newfloat{listing}{tb}{lst}{}
\floatname{listing}{Listing}
\title{Adaptive Anytime Multi-Agent Path Finding Using\\Bandit-Based Large Neighborhood Search}
\author {
    Thomy Phan,
    Taoan Huang,
    Bistra Dilkina,
    Sven Koenig
}
\affiliations {
    University of Southern California, USA\\
    \{thomy.phan, taoanhua, dilkina, skoenig\}@usc.edu
}

\begin{document}

\maketitle

\begin{abstract}
Anytime \emph{multi-agent path finding (MAPF)} is a promising approach to scalable path optimization in large-scale multi-agent systems. State-of-the-art anytime MAPF is based on \emph{Large Neighborhood Search (LNS)}, where a fast initial solution is iteratively optimized by destroying and repairing a fixed number of parts, i.e., the neighborhood of the solution, using randomized destroy heuristics and prioritized planning.
Despite their recent success in various MAPF instances, current LNS-based approaches lack exploration and flexibility due to greedy optimization with a fixed neighborhood size which can lead to low-quality solutions in general. So far, these limitations have been addressed with extensive prior effort in tuning or offline machine learning beyond actual planning.
In this paper, we focus on online learning in LNS and propose \emph{Bandit-based Adaptive LArge Neighborhood search Combined with Exploration (BALANCE)}. BALANCE uses a bi-level multi-armed bandit scheme to adapt the selection of destroy heuristics and neighborhood sizes on the fly during search. We evaluate BALANCE on multiple maps from the MAPF benchmark set and empirically demonstrate performance improvements of at least 50\% compared to state-of-the-art anytime MAPF in large-scale scenarios. We find that Thompson Sampling performs particularly well compared to alternative multi-armed bandit algorithms.
\end{abstract}

\section{Introduction}

A wide range of real-world applications like goods transportation in warehouses,  search and rescue missions,  and traffic management can be formulated as \emph{Multi-Agent Path Finding (MAPF)} problem,  where the goal is to find collision-free paths for multiple agents with each having an assigned start and goal location. Finding optimal solutions, w.r.t.  minimal flowtime or makespan is NP-hard,  which limits scalability of most state-of-the-art MAPF solvers \cite{ratner1986finding,yu2013structure,sharon2012conflict}.

\emph{Anytime MAPF} based on \emph{Large Neighborhood Search (LNS)} is a popular approach to finding fast and near-optimal solutions to the MAPF problem within a fixed time budget \cite{li2021anytime}. Given an initial feasible solution and a set of destroy heuristics, LNS iteratively destroys and replans so-called neighborhoods of the solution, i.e., a fixed number of paths, until the time budget runs out. MAPF-LNS represents the current state-of-the-art in anytime MAPF and has been experimentally shown to scale up to large-scale scenarios with hundreds of agents \cite{li2021anytime}. Due to its increasing popularity, several extensions have been recently proposed like fast local repairing, integration of primal heuristics, or machine learning-guided neighborhood selection \cite{HuangAAAI22,li2022lns2,LamICAPS23}.

However, MAPF-LNS and its variants currently suffer from two limitations that can lead to low-quality solutions in general:
\begin{enumerate}
\item The neighborhood size is typically fixed, which limits the flexibility of the optimization process, thus possibly affecting the solution quality, especially for a large number of agents \cite{li2021anytime}. Therefore, prior tuning is required -- in addition to the actual LNS procedure -- to obtain good solutions.
\item Roulette wheel selection is commonly used to execute and adapt the destroy heuristic selection to determine the neighborhood \cite{mara2022survey,li2021anytime}. During optimization, roulette wheel selection could greedily converge to poor choices due to the lack of exploration. Offline machine learning can guide the selection with solution score prediction but requires sufficient data acquisition and feature engineering \cite{HuangAAAI22}.
\end{enumerate}

In this paper, we address these limitations by proposing \emph{Bandit-based Adaptive LArge Neighborhood search Combined with Exploration (BALANCE)}.
BALANCE uses a bi-level multi-armed bandit scheme to adapt the selection of destroy heuristics and neighborhood sizes on the fly during search.
Our contributions are as follows:
\begin{itemize}
\item We formulate BALANCE as a simple but effective MAPF-LNS framework with an adaptive selection of destroy heuristics and neighborhood sizes during search.
\item We propose and discuss three concrete instantiations of BALANCE based on roulette wheel selection, UCB1, and Thompson Sampling, respectively.
\item We evaluate BALANCE on multiple maps from the MAPF benchmark set and empirically demonstrate cost improvements of at least 50\% compared to state-of-the-art anytime MAPF in large-scale scenarios. We find that Thompson Sampling performs particularly well compared to alternative multi-armed bandit algorithms.
\end{itemize}

\section{Background}\label{sec:background}

\subsection{Multi-Agent Path Finding (MAPF)}

We focus on \emph{maps} as undirected unweighted \emph{graphs} $G = \langle \mathcal{V}, \mathcal{E} \rangle$, where vertex set $\mathcal{V}$ contains all possible locations and edge set $\mathcal{E}$ contains all possible transitions or movements between adjacent locations. An \emph{instance} $I$ consists of a map $G$ and a set of \emph{agents} $\mathcal{A} = \{a_1, ..., a_m\}$ with each agent $a_i$ having a \emph{start location} $s_i \in \mathcal{V}$ and a \emph{goal location} $g_i \in \mathcal{V}$.

MAPF aims to find a collision-free plan for all agents. A \emph{plan} $P = \{ p_1, ..., p_m \}$ consists of individual paths $p_i = \langle p_{i,1}, ..., p_{i,l(p_i)} \rangle$ per agent $a_i$, where $\langle p_{i,t}, p_{i,t+1} \rangle = \langle p_{i,t+1}, p_{i,t} \rangle \in \mathcal{E}$, $p_{i,1} = s_i$, $p_{i,l(p_i)} = g_i$, and $l(p_i)$ is the \emph{length} or \emph{travel distance} of path $p_i$. The \emph{delay} $\textit{del}(p_i)$ of path $p_i$ is defined by the difference of path length $l(p_i)$ and the length of the shortest path from $s_i$ to $g_i$ w.r.t. map $G$.

In this paper, we consider \emph{vertex conflicts} $\langle a_i, a_j, v, t \rangle$ that occur when two agents $a_i$ and $a_j$ occupy the same location $v \in \mathcal{V}$ at time step $t$ and \emph{edge conflicts} $\langle a_i, a_j, u, v, t \rangle$ that occur when two agents $a_i$ and $a_j$ traverse the same edge $\langle u, v \rangle \in \mathcal{E}$ in opposite directions at time step $t$ \cite{stern2019multi}. A plan $P$ is a \emph{solution}, i.e., \emph{feasible}, when it does not have any vertex or edge conflicts. Our goal is to find a solution that minimizes the \emph{flowtime} $\sum_{p \in P} l(p)$ which is equivalent to minimizing the \emph{sum of delays} $\sum_{p \in P} \textit{del}(p)$. We use the sum of delays or \emph{(total) cost} $c(P) = \sum_{p \in P} \textit{del}(p)$ as the primary performance measure in our evaluations.

\subsection{Anytime MAPF with LNS}

\emph{Anytime MAPF} searches for solutions within a given time budget. The solution quality monotonically improves with increasing time budget \cite{cohen2018anytime,li2021anytime}.

\emph{MAPF-LNS} based on \emph{Large Neighborhood Search (LNS)} is the current state-of-the-art approach to anytime MAPF and is shown to scale up to large-scale scenarios with hundreds of agents \cite{HuangAAAI22,li2021anytime}. Starting with an initial feasible plan $P$, e.g., found via \emph{prioritized planning (PP)} from \cite{silver2005cooperative}, MAPF-LNS iteratively modifies $P$ by destroying $N < m$ paths, i.e., the \emph{neighborhood} $P^{-} \subset P$. The destroyed neighborhood is then repaired or replanned using PP to quickly generate a new solution $P^{+}$. If the new cost $c(P^{+})$ is lower than the previous cost $c(P)$, then $P$ is replaced by $P^{+}$, and the search continues until the time budget runs out. The result of MAPF-LNS is the last accepted solution $P$ with the lowest cost so far.

MAPF-LNS uses a set $\mathcal{H}$ of three \emph{destroy heuristics} $H \in \mathcal{H}$, namely a \emph{random uniform selection} of $N$ paths, an \emph{agent-based heuristic}, and a \emph{map-based heuristic} \cite{li2021anytime}. The agent-based heuristic generates the neighborhood, including the path of agent $a_i$ with the current maximum delay and other paths (determined via random walks) that prevent $a_i$ from achieving a lower delay. The map-based heuristic randomly chooses a vertex $v \in \mathcal{V}$ with a degree greater than 2 and generates a neighborhood of paths containing $v$.

MAPF-LNS uses a \emph{selection algorithm} $\pi$ like roulette wheel selection to choose destroy heuristics $H \in \mathcal{H}$ by maintaining updatable \emph{weights} or some \emph{statistics} for all destroy heuristics \cite{ropke2006adaptive,li2021anytime}. All weights or statistics used by $\pi$ to select a destroy heuristic $H$ are denoted by $\Delta$, which could represent, e.g., the average cost improvement or the selection count per destroy heuristic $H$. The statistics $\Delta$ will be further explained in Section \ref{subsec:balance_instantiations} as the concrete definition depends on $\pi$.

\subsection{Multi-Armed Bandits}
\emph{Multi-armed bandits (MABs)} or simply bandits are fundamental decision-making problems, where an \emph{MAB or selection algorithm} $\pi$ repeatedly chooses an \emph{arm} $k$ among a given set of arms or \emph{options} $\{1, ..., K\}$ to maximize an expected \emph{reward} of a stochastic reward function $\mathcal{R}(k) := X_{k}$, where $X_{k}$ is a random variable with an unknown distribution $f_{X_{k}}$. To solve an MAB, one has to determine an \emph{optimal arm} $k^{*}$, which maximizes the expected reward $\mathbb{E}\big[X_{k}\big]$. The MAB algorithm $\pi$ has to balance between sufficiently exploring all arms $k$ to accurately estimate $\mathbb{E}\big[X_{k}\big]$ via statistics $\Delta$ and to exploit its current estimates by greedily selecting the arm $k$ with the currently highest estimate of $\mathbb{E}\big[X_{k}\big]$. This is known as the \emph{exploration-exploitation dilemma}, where exploration can find arms with higher rewards but requires more time for trying them out, while exploitation can lead to fast convergence but possibly gets stuck in a poor local optimum. In this paper, we will cover \emph{roulette wheel selection}, \emph{UCB1}, and \emph{Thompson Sampling} as concrete MAB algorithms and further explain them in Section \ref{subsec:balance_instantiations}.

\section{Related Work}

\subsection{Multi-Armed Bandits for LNS}
In recent years, MABs have been used as adaptive meta-controllers to tune learning and optimization algorithms on the fly \cite{schaul2019adapting,badia2020agent57,hendel2022adaptive}. Besides roulette wheel selection, UCB1 and $\epsilon$-greedy are commonly used for destroy heuristic selection in LNS in the context of mixed integer programming, vehicle routing, and scheduling problems with fixed neighborhood sizes \cite{Chen2016AMB,Chen2018ARL,chmiela2023online}. \cite{hendel2022adaptive} adapts the neighborhood size for mixed integer programming using a mutation-based approach inspired by evolutionary algorithms \cite{rothberg2007evolutionary}. Most works use rather complex rewards that are composed of multiple weighted terms with several tunable hyperparameters.
We focus on \emph{MAPF problems} and propose a \emph{bi-level MAB scheme} to adapt the selection of destroy heuristics and neighborhood sizes, which is simple to use without any additional mechanisms like mutation. Our approach uses the \emph{cost improvement} as a reward, which simply represents the cost difference between two solutions w.r.t. the original objective of MAPF without depending on any additional weighted term that requires prior tuning. To the best of our knowledge, our work first effectively applies \emph{Thompson Sampling} to anytime MAPF in addition to more common MAB algorithms like UCB1 and roulette wheel selection.

\subsection{Multi-Armed Bandits in Anytime Planning}

MABs are popular in anytime planning algorithms, especially in single-agent Monte Carlo planning \cite{kocsis2006bandit,silver2010monte}. \emph{Monte-Carlo Tree Search (MCTS)} is the state-of-the-art framework of current Monte Carlo planning algorithms which uses MABs to traverse a search tree within a limited time budget \cite{kocsis2006bandit,silver2010monte}. UCB1 is most commonly used, but Thompson Sampling has also gained attention in the last few years due to its effectiveness in domains of high uncertainty \cite{bai2013bayesian,bai2014thompson,phan2019memory,phan2019adaptive}.
As MABs have been shown to converge to good decisions within short-time budgets, we use MABs in our adaptive \emph{multi-agent path finding} setting. Inspired by the latest progress in Monte Carlo planning \cite{swiechowski2023monte}, we intend to employ more sophisticated MAB algorithms like Thompson Sampling to anytime MAPF to improve exploration and performance.

\subsection{Machine Learning in Anytime MAPF}

Machine learning has been used in MAPF to directly learn collision-free path finding, to guide node selection in search trees, or to select appropriate MAPF algorithms for certain maps \cite{sartoretti2019primal,kaduri2020algorithm,huang2021learning}. MAPF-ML-LNS is an anytime MAPF approach that extends MAPF-LNS with a learned score predictor for neighborhood selection as well as a random uniform selection of the neighborhood size $N$. The predictor is trained offline on pre-collected data from previous MAPF runs \cite{HuangAAAI22}. The score predictor generalizes to some degree but is fixed after training; therefore, not being able to adapt during search, which limits flexibility. MAPF-ML-LNS depends on extensive prior effort like data acquisition, model training, and feature engineering for meaningful score learning.
We propose an \emph{online learning} approach to adaptive MAPF-LNS using MABs. The MABs can be trained on the fly with data \emph{directly} obtained from the LNS without any prior data acquisition. Since MABs only learn from \emph{scalar rewards}, there is no need for expensive feature engineering, simplifying our approach and easing application to other domains.

\section{Bandit-Based Adaptive MAPF-LNS}\label{sec:balance}

We now introduce \emph{Bandit-based Adaptive LArge Neighborhood search Combined with Exploration (BALANCE)} as a simple but effective LNS framework for adaptive MAPF.

\subsection{Formulation}

\begin{figure}
  \centering
  \includegraphics[width=0.35\textwidth]{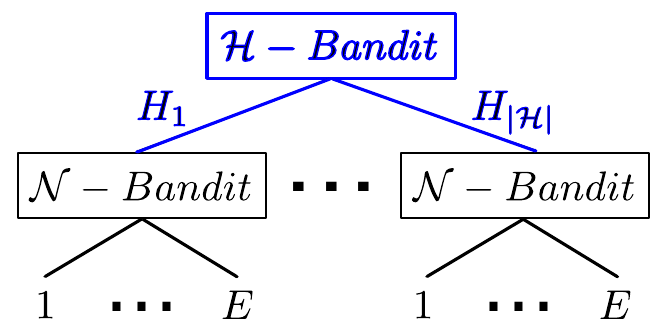}
  \caption{Bi-level multi-armed bandit scheme of BALANCE. The top-level $\mathcal{H}$-Bandit selects a destroy heuristic $H \in \mathcal{H}$. Each bottom-level $\mathcal{N}$-Bandit corresponds to a destroy heuristic choice and selects an exponent $e \in \mathcal{N} = \{1,...,E\}$ to determine the neighborhood size $N = 2^{e}$.}
  \label{fig:balance_scheme}
\end{figure}

BALANCE uses a bi-level MAB scheme to adapt the selection of destroy heuristics and neighborhood sizes on the fly during search.
The first level consists of a single MAB, called $\mathcal{H}$\emph{-Bandit} with $K = |\mathcal{H}|$ arms, which selects a destroy heuristic $H \in \mathcal{H}$. The second level consists of $|\mathcal{H}|$ so-called $\mathcal{N}$\emph{-Bandits} with $K = E$ arms. Each $\mathcal{N}$-Bandit conditions on a destroy heuristic choice $H \in \mathcal{H}$ and determines the corresponding neighborhood size $N \in \mathcal{N} = \{2^e | e \in \{1,...,E\}\}$\footnote{The set of neighborhood size options $\mathcal{N}$ can be defined arbitrarily. For simplicity, we focus on sets consisting of powers of two.} based on an exponent selection $e \in \{1,...,E\}$. The bi-level MAB scheme is shown in Figure \ref{fig:balance_scheme}.

BALANCE first selects a destroy heuristic $H$ with the top-level $\mathcal{H}$-Bandit based on its current statistics $\Delta_{\mathcal{H}}$. The selected destroy heuristic $H$ determines the corresponding bottom-level $\mathcal{N}$-Bandit, which is used to select an exponent $e$ based on its current conditional statistics $\Delta^{\mathcal{N}}_{H}$. The neighborhood size is then determined by $N = 2^{e}$.
After evaluating the total cost $c(P^{+})$ of the new solution $P^{+}$, i.e. the sum of delays, the statistics of the top-level $\mathcal{H}$-Bandit and corresponding bottom-level $\mathcal{N}$-Bandit are updated incrementally. The \emph{MAB reward} $x_k = x_{\langle H,N\rangle} = \textit{max}\{0, c(P) - c(P^{+})\}$ for the update is defined by the \emph{cost improvement} of the new solution $P^{+}$ compared to the previous one $P$ \cite{li2021anytime}.

The full formulation of BALANCE is provided in Algorithm \ref{algorithm:BALANCE}, where $I$ represents the instance to be solved, $\pi$ represents the MAB algorithm for the bi-level scheme, and $E$ represents the number of neighborhood size options.

\begin{algorithm}
\caption{BALANCE as MAPF-LNS Framework}\label{algorithm:BALANCE}
\begin{algorithmic}[1]
\Procedure{$\textit{BALANCE}(I, \pi, E)$}{}
\State $P = \{ p_1, ..., p_m \} \leftarrow \textit{RunInitialSolver(I)}$
\State Initialize statistics $\Delta$ of MAB algorithm $\pi$
\While{runtime limit not exceeded}
\State $\langle H, N \rangle \sim \textit{BiLevelBanditSelection}(\pi, \Delta, E)$
\State $A \sim \textit{SampleNeighborhood}(I, H, N)$
\State $P^{-} \leftarrow \{p_i | a_i \in A\}$
\State $P^{+} \leftarrow \textit{RunRepairSolver}(I, A, P \backslash P^{-})$
\State $x_{\langle H,N\rangle} \leftarrow \textit{max}\{0, c(P) - c(P^{+})\}$
\If{$x_{\langle H,N\rangle} > 0$}\Comment{check cost improvement}
\State $P \leftarrow (P \backslash P^{-}) \cup P^{+}$
\EndIf
\State $\textit{UpdateBandits}(\Delta, H, N, x_{\langle H,N\rangle})$
\EndWhile
\Return $P$
\EndProcedure
\end{algorithmic}
\end{algorithm}

\subsection{Instantiations}\label{subsec:balance_instantiations}

In the following, we describe three concrete MAB algorithms $\pi$ to implement the bi-level scheme in Figure \ref{fig:balance_scheme}. The definition of the statistics $\Delta$ depends on the MAB algorithm.

\subsubsection{Roulette Wheel Selection}
$\pi$ selects an arm $k$ with a probability of $\frac{w_k}{\sum^{K}_{j=1}w_j}$, where $w_k = \sum^{T_k}_{c=1} x^{(c)}_{k}$ is the \emph{sum of rewards} or \emph{weight} and $T_k$ is the \emph{selection count} of arm $k$. Statistics $\Delta$ consists of all weights $w_k$, which can be updated incrementally after each iteration \cite{Goldberg1988GeneticAI}.

\subsubsection{UCB1}

$\pi$ selects arms by maximizing the \emph{upper confidence bound} of rewards $\textit{UCB1}(k) = \overline{x}_k + \xi \sqrt{\frac{\textit{log}(T)}{T_{k}}}$, where $\overline{x}_k$ is the \emph{average reward} of arm $k$, $\xi$ is an exploration constant, $T$ is the total number of arm selections, and $T_{k}$ is the selection count of arm $k$. The second term represents the \emph{exploration bonus}, which becomes smaller with increasing $T_{k}$ \cite{auer2002finite}. Statistics $\Delta$ consists of all average rewards $\overline{x}_k$ and selection counts $T_{k}$.

\subsubsection{Thompson Sampling}
$\pi$ uses a Bayesian approach to balance between exploration and exploitation of arms \cite{thompson1933likelihood}.
We focus on a generalized variant of Thompson Sampling, which works for arbitrary reward distributions $f_{X_{k}}$ by assuming that $X_k$ follows a Normal distribution $\mathcal{N}(\mu_k, \frac{1}{\tau_k})$ with unknown mean $\mu_k$ and precision $\tau_k = \frac{1}{\sigma^2_k}$, where $\sigma^2_k$ is the variance \cite{bai2013bayesian,bai2014thompson}. $\langle\mu_k, \tau_k\rangle$ follows a Normal Gamma distribution $\mathcal{NG}(\mu_{0},\lambda_k,\alpha_k,\beta_k)$ with $\lambda_k > 0$, $\alpha_k \geq 1$, and $\beta_k \geq 0$. The distribution over $\tau_k$ is a Gamma distribution $\tau_k \sim \textit{Gamma}(\alpha_k,\beta_k)$ and the conditional distribution over $\mu_k$ given $\tau_k$ is a Normal distribution $\mu_k \sim \mathcal{N}(\mu_{0},\frac{1}{\lambda_k\tau_k})$. Given a \emph{prior distribution} $P(\theta) = \mathcal{NG}(\mu_{0},\lambda_{0},\alpha_{0},\beta_{0})$ and $n$ observed rewards $D_k = \{x^{(1)}_{1},...,x^{(T_k)}_{k}\}$, the \emph{posterior distribution} is defined by $P(\theta|D_k) = \mathcal{NG}(\mu_{k,1},\lambda_{k,1},\alpha_{k,1},\beta_{k,1})$, where $\mu_{k,1} = \frac{\lambda_{0}\mu_{0} + T_k \overline{x}_k}{\lambda_{0}+T_k}$,
$\lambda_{k,1} = \lambda_{0}+T_k$, $\alpha_{k,1} = \alpha_{0}+\frac{T_k}{2}$, and $\beta_{k,1} = \beta_{0}+\frac{1}{2}(T_k \sigma^{2}_k + \frac{\lambda_{0}T_k(\overline{x}_k - \mu_{0})^2}{\lambda_{0}+T_k})$. $\overline{x}_k$ is the observed average reward in $D_k$ and $\sigma^{2}_k = \frac{1}{T_k}\sum_{c = 1}^{T_k} (x^{(c)}_{k} - \overline{x}_k)^2$ is the variance. The posterior is inferred for each arm $k$ to sample an estimate $\mu_{k}$ of the expected reward $\mathbb{E}\big[X_{k}\big]$. The arm with the highest estimate is selected. 
Statistics $\Delta$ consists of all average rewards $\overline{x}_k$, \emph{average of squared rewards} $\frac{1}{T_k}\sum^{T_k}_{c=1} (x^{(c)}_{k})^2$, and selection counts $T_{k}$.

\subsection{Conceptual Discussion}\label{subsec:conceptual_discussion}

As MAB algorithms balance between exploration and exploitation to quickly find optimal choices, we believe that they are naturally suited to enhance MAPF-LNS with self-adaptive capabilities. According to previous works on MAB-based tree search, BALANCE can provably converge to an optimal destroy heuristic and neighborhood size choice with sufficient exploration if there is a \emph{stationary optimum} \cite{kocsis2006bandit,bai2014thompson}. Otherwise, non-stationary MAB techniques are required, which we defer to as future work \cite{garivier2008upper}. Depending on the choice of $E$, BALANCE maintains $E + 1$ MABs in total. Since $\Delta$ can be updated incrementally for any quantity like arm selection counts $T_k$ or average rewards $\overline{x}_k$, the bi-level MAB scheme can be updated in constant time thus introducing negligible overhead to the LNS (as replanning of neighborhoods requires significantly more compute).

Roulette wheel selection is the simplest method to implement because it only uses the weights $w_k$ as the sum of rewards. However, it could lack exploration in the long run since arms with small weights are likely to be neglected or forgotten over time. UCB1 accommodates this issue by introducing an exploration bonus that explicitly considers the selection count of arms $T_k$. Arms that are selected less over time will have a larger exploration bonus and are therefore more incentivized for selection, depending on the choice of exploration constant $\xi$. Thompson Sampling is a randomized algorithm whose initial exploration depends on prior parameters, i.e., $\mu_{0}$, $\lambda_0$, $\alpha_0$, and $\beta_0$ thus being more complex than the other MAB approaches. However, previous works report that using prior distributions that are close to a \emph{uniform distribution} is sufficient in most cases without requiring extensive tuning \cite{bai2013bayesian,bai2014thompson}.

Adaptation in MAPF-LNS can be regarded as \emph{stochastic optimization problem}, since all destroy heuristics defined by \cite{li2021anytime} are randomized. Therefore, uncertainty-based methods like Thompson Sampling seem promising for this setting as reported in \cite{chapelle2011empirical,kaufmann2012thompson,bai2014thompson}.

Alternatively to the proposed bi-level MAB scheme, a single MAB can be employed to directly search the joint arm space of $\mathcal{H} \times \mathcal{N}$. While this approach would basically solve the same problem, the joint arm space scales quadratically, which could lead to low-quality solutions, if the time budget is very restricted. The bi-level scheme mitigates the scalability issue by first selecting a destroy heuristic $H$ (Section \ref{subsec:results_exploration} indicates that performance is more sensitive to $H$) before deciding on the neighborhood size $N$ (whose quality depends on the choice of $H$).

\section{Experiments}\label{sec:experiments}

\subsection{Setup}

\subsubsection{Maps}

We evaluate BALANCE on five maps from the MAPF benchmark set of \cite{stern2019multi}, namely (1) a \texttt{random} map (\emph{random-32-32-10}), (2) a \texttt{warehouse} map (\emph{warehouse-10-20-10-2-1}), (3) two \texttt{game} maps \emph{ost003d} and (4) \emph{den520d} as well as (5) a \texttt{city} map (\emph{Paris\_1\_256}). All maps have different sizes and structures and are the same as used in \cite{HuangAAAI22} for comparability with state-of-the-art anytime MAPF as presented below. We conduct all experiments on the available 25 random scenarios per map.

\subsubsection{Anytime MAPF Algorithms} We implemented\footnote{Code available at \underline{\url{github.com/thomyphan/anytime-mapf}}.} different variants of BALANCE using Thompson Sampling (with $\mu_{0} = 0$, $\lambda_0 = 0.01$, $\alpha_0 = 1$, $\beta_0 = 100$), UCB1 (with $\xi = 1,000$), and roulette wheel selection. Each BALANCE variant is denoted by \textit{BALANCE (X)}, where \emph{X} is the concrete MAB algorithm (or just random uniform sampling) used for our bi-level scheme in Figure \ref{fig:balance_scheme}. Unless stated otherwise, we always use the $|\mathcal{H}| = 3$ destroy heuristics from \cite{li2021anytime} and set $E = 5$ such that the neighborhood size is chosen from $\mathcal{N} = \{2, 4, 8, 16, 32\}$\footnote{Even though previous works \cite{li2021anytime,li2022lns2} already indicate good values for fixed neighborhood sizes $N$, we keep optimizing our MABs on a broader set of options to confirm convergence to adequate choices without assuming any prior knowledge.}. Our BALANCE implementation is based on the public code of \cite{li2022lns2} and uses its default configuration unless stated otherwise.

We determine the \emph{Empirically Best Choice}, where we run a grid search over all $|\mathcal{H}| = 3$ destroy heuristics and $|\mathcal{N}| = E = 5$ neighborhood size options $\mathcal{N} = \{2, 4, 8, 16, 32\}$ to compare with a pre-tuned LNS without any adaptation.

To directly compare BALANCE with MAPF-LNS and MAPF-ML-LNS, as state-of-the-art approaches, we take the performance values reported in \cite{HuangAAAI22}, running our experiments on the same hardware specification. We also compare with a single-MAB approach that optimizes over the $\mathcal{H} \times \mathcal{N}$ \emph{Joint Arm Space} using Thompson Sampling.

\subsubsection{Compute Infrastructure}

All experiments were run on a x86\_64 GNU/Linux (Ubuntu 18.04.5 LTS) machine with i7 @ 2.4 GHz CPU and 16 GB RAM, as in \cite{HuangAAAI22}.

\subsection{Experiment -- BALANCE Convergence}

\subsubsection{Setting}

To assess convergence w.r.t. time budget, we run \emph{BALANCE (Thompson)}, \emph{BALANCE (UCB1)}, \emph{BALANCE (Roulette)}, and \emph{BALANCE (Random)} on the \texttt{random} and \texttt{city} map with $m = 200$ and $350$ agents respectively.%We additionally determine the \emph{empirically best choice}, where we run a grid search over all $|\mathcal{H}| = 3$ destroy heuristics and $|\mathcal{N}| = E = 5$ neighborhood size options $\mathcal{N} = \{2, 4, 8, 16, 32\}$ to compare with a pre-tuned LNS without any online adaptation.

\subsubsection{Results}

\begin{figure}
	\centering
	\includegraphics[width=0.47\textwidth]{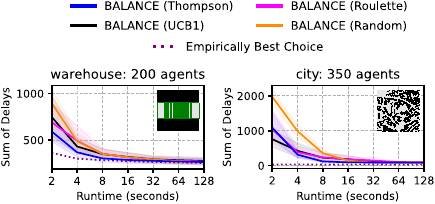}
     \caption{Sum of delays for different BALANCE variants with different time budgets compared with the respective empirically best choices. Shaded areas show the 95\% confidence interval. The legend at the top applies across all plots.}
     \label{fig:balance_results_runtime}
\end{figure}

The results are shown in Figure \ref{fig:balance_results_runtime}. With increasing time budget, all BALANCE variants converge to an average sum of delays close to the empirically best choice. All MAB-enhanced variants converge faster than \emph{BALANCE (Random)}. \emph{BALANCE (Thompson)} performs best in both maps, especially when the time budget is low.

\subsubsection{Discussion} 

The results show that any version of BALANCE is able to perform well with an increasing time budget. Given a sufficient time budget, all versions are able to keep up with the empirically best choice through online learning without running a prior grid search that requires roughly $|\mathcal{H}|\cdot|\mathcal{N}| = 3E = 15$ times the compute of any BALANCE variant in total. Thompson Sampling performs particularly well, presumably due to the inherent uncertainty exhibited by the randomized destroy heuristics.
%- Random might be a reasonable choice if no further knowledge available.\\

\subsection{Experiment -- BALANCE Exploration}\label{subsec:results_exploration}

\subsubsection{Setting}

Next, we evaluate the explorative behavior of \emph{BALANCE (Thompson)}, \emph{BALANCE (UCB1)}, and \emph{BALANCE (Roulette)} on the \texttt{random}, \texttt{ost003d}, and \texttt{city} map after 128 seconds of LNS runtime. We also evaluate the progress of MAB choice over time for \emph{BALANCE (Thompson)} and \emph{BALANCE (Roulette)} in the \texttt{ost003d} map.

\subsubsection{Results}

\begin{figure*}
	\centering
	\includegraphics[width=0.83\textwidth]{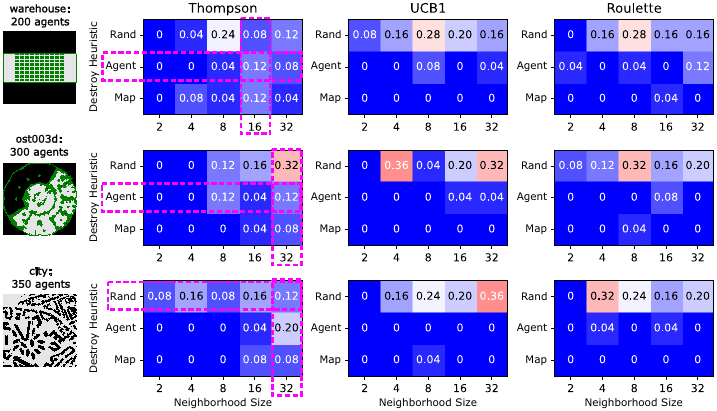}
     \caption{Relative frequencies of selected destroy heuristic and neighborhood size combinations $\langle H, N \rangle$ per BALANCE variant after 128 seconds of planning. Magenta dashed boxes indicate the empirically best destroy heuristic and neighborhood size.}
     \label{fig:balance_results_heatmaps}
\end{figure*}

The final relative frequencies of MAB choices are displayed as heatmaps in Figure \ref{fig:balance_results_heatmaps}. The empirically best destroy heuristics and neighborhood sizes are highlighted by magenta dashed boxes. \emph{BALANCE (UCB1)} and \emph{BALANCE (Roulette)} strongly prefer the random destroy heuristic, while the preferred neighborhood size depends on the actual map. \emph{BALANCE (Thompson)} also prefers the random destroy heuristic to some degree but still explores other heuristics, mainly with neighborhood sizes $N \geq 8$. Compared to the other variants, \emph{BALANCE (Thompson)} explores more regions where either the destroy heuristic $H$ or the neighborhood size $N$ is empirically best, at least.

%To quantify the exploration relative to the empirically best choice, we regard $H$ and $N$ as \emph{stochastic events} and classify them either as empirically best ($b$) or not ($\neg b$). The \emph{normalized joint entropy} $\eta = -\frac{1}{\eta_{\textit{uniform}}}\sum_{H \in \{b, \neg b\}} \sum_{N \in \{b, \neg b\}}\textit{Pr}(H,N)\textit{log}(\textit{Pr}(H,N))$ of both events is shown in Figure \ref{fig:balance_entropy} for each MAB, where Thompson Sampling is shown to explore the event combinations $\{b, \neg b\}$ of $H$ and $N$ the most -- even more than random uniform in \texttt{ost003d} and \texttt{city}.

\begin{figure}
	\centering
	\includegraphics[width=0.44\textwidth]{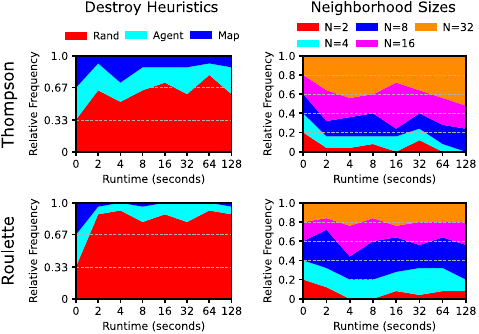}
     \caption{MAB choices over time for \texttt{ost003d}.}
     \label{fig:MAB_progress}
\end{figure}

Figure \ref{fig:MAB_progress} shows the average progress of the chosen destroy heuristic $H$ and neighborhood size $N$ during search for Thompson Sampling and Roulette in the \texttt{ost003d} map. While Roulette quickly converges to the random heuristic, Thompson Sampling
adapts its preferences through continuous exploration. Thompson Sampling mostly prefers the largest neighborhood size $N = 32$ over time, whereas Roulette almost uniformly chooses $N \in \{4, 8, 16, 32\}$ over time with a slight preference toward $N = 8$.

\subsubsection{Discussion}

None of the BALANCE variants clearly converges to the empirically best choice, which could be due to a short time budget, marginal improvement over time, or potential non-stationarity of the actual optimal choice. Nevertheless, Figure \ref{fig:balance_results_heatmaps} suggests that Thompson Sampling performs more focused exploration than any other MAB.

\subsection{Experiment -- Neighborhood Size Options}\label{subsec:results_neighborhood}

\subsubsection{Setting}

We run \emph{BALANCE (Thompson)}, \emph{BALANCE (UCB1)}, \emph{BALANCE (Roulette)}, and \emph{BALANCE (Random)} with different neighborhood size options by varying $E$, i.e., the number of exponents $e$ to determine the neighborhood size $N = 2^{e}$. The same maps and number of agents $m$ as above are used with a time budget of 128 seconds. We additionally evaluate with a doubled number of agents per map.

\subsubsection{Results}

\begin{figure*}
	\centering
	\includegraphics[width=0.95\textwidth]{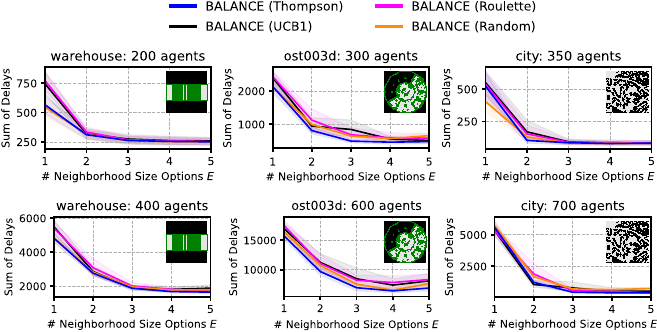}
     \caption{Sum of delays for different BALANCE variants with different neighborhood size options $E$ and numbers of agents $m$. The time budget is 128 seconds. Shaded areas show the 95\% confidence interval. The legend at the top applies across all plots.}
     \label{fig:balance_results_neighborhoods}
\end{figure*}

The results are shown in Figure \ref{fig:balance_results_neighborhoods}. All approaches significantly improve when the number of options is increased to $E = 3$ with marginal to no improvement afterward. \emph{BALANCE (Thompson)} and \emph{BALANCE (Random)} benefit the most from the increase of $E$ except in the \texttt{city} map with 700 agents, where \emph{BALANCE (UCB1)} keeps up with \emph{BALANCE (Thompson)}.

\subsubsection{Discussion}

Since Thompson Sampling and random uniform explore more than UCB1 and Roulette, they can better leverage the neighborhood size options. The results indicate that neighborhood size adaptation and the sufficient availability of options can significantly affect performance. However, the neighborhood size also affects the amount of compute for replanning, which explains why \emph{BALANCE (Random)} performs worse in \texttt{ost003d} when $E = 5$.

\subsection{Experiment -- State-of-the-Art Comparison}\label{subsec:star_results}

\subsubsection{Setting}

We run \emph{BALANCE (Thompson)}, \emph{BALANCE (UCB1)}, \emph{BALANCE (Roulette)}, and \emph{Joint Arm Space (Thompson)} on the \texttt{random}, \texttt{warehouse}, \texttt{ost003d}, \texttt{den520d}, and \texttt{city} map with different numbers of agents $m$. For direct comparability with MAPF-LNS and MAPF-ML-LNS, we set the time budget to 60 seconds \cite{HuangAAAI22}. Since no error or deviation bars are reported in \cite{HuangAAAI22}, we only show the average performance of MAPF-LNS and MAPF-ML-LNS as dashed lines.

\subsubsection{Results}

\begin{figure*}
	\centering
	\includegraphics[width=0.95\textwidth]{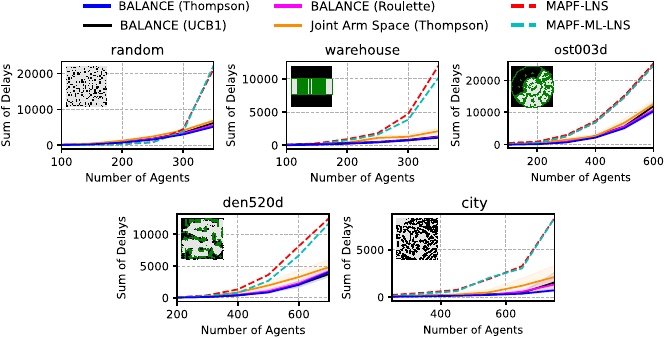}
     \caption{Sum of delays for different variants of BALANCE compared with state-of-the-art anytime MAPF-LNS and MAPF-ML-LNS for different numbers of agents $m$. The performance values of MAPF-LNS and MAPF-ML-LNS are taken from \cite{HuangAAAI22} without any error or deviation bars. Our experiments are run on the same hardware specification with a time budget of 60 seconds. Shaded areas show the 95\% confidence interval. The legend at the top applies across all plots.}
     \label{fig:balance_results_star}
\end{figure*}

The results are shown in Figure \ref{fig:balance_results_star}. All BALANCE variants and \emph{Joint Arm Space (Thompson)} significantly outperform MAPF-LNS and MAPF-ML-LNS by at least 50\% when $m \geq 350$. In the \texttt{random}, \texttt{ost003d}, and \texttt{city} map, \emph{BALANCE (Thompson)} slightly outperforms the other BALANCE variants. \emph{Joint Arm Space (Thompson)} is consistently outperformed by the BALANCE variants.

\subsubsection{Discussion}

The experiment demonstrates that BALANCE effectively mitigates the limitations of state-of-the-art anytime MAPF regarding fixed neighborhood sizes and the lack of exploration in roulette wheel selection, especially in instances with a large number of agents $m$. While Thompson Sampling seemingly performs best in most cases, using BALANCE with any MAB algorithm is generally beneficial to improve performance. As discussed in Section \ref{subsec:conceptual_discussion}, our bi-level MAB scheme can outperform joint arm space alternatives when the time budget is very restricted due to meaningful decomposition, which is confirmed in all tested maps. However, since \emph{Joint Arm Space (Thompson)} also outperforms the state-of-the-art, we suggest that bandit-based adaptation in MAPF-LNS is generally promising.

\section{Conclusion}

We presented BALANCE, an LNS framework using a bi-level multi-armed bandit scheme to adapt the selection of destroy heuristics and neighborhood sizes during search.

Our experiments show that BALANCE offers a simple but effective framework for adaptive anytime MAPF, which is able to significantly outperform state-of-the-art anytime MAPF without requiring extensive prior efforts like neighborhood size tuning, data acquisition, or feature engineering. Sufficient availability of neighborhood size options is important to provide enough room for adaptation at the potential cost of runtime due to increasing replanning effort. Thompson Sampling is a promising choice for most scenarios due to the inherent uncertainty of the randomized destroy heuristics and its ability to explore promising choices.

Future work includes the investigation of non-stationary MAB approaches and online learnable destroy heuristics.

\section*{Acknowledgements}
The research at the University of Southern California was supported by the National Science Foundation (NSF) under grant numbers 1817189, 1837779, 1935712, 2121028, 2112533, and 2321786 as well as a gift from
Amazon Robotics. The views and conclusions contained in this document are those of the authors and should not be interpreted as representing the official policies, either expressed or implied, of the sponsoring organizations, agencies, or the U.S.  government.

\bibliography{aaai24}

\begin{thebibliography}{35}
\providecommand{\natexlab}[1]{#1}

\bibitem[{Auer, Cesa-Bianchi, and Fischer(2002)}]{auer2002finite}
Auer, P.; Cesa-Bianchi, N.; and Fischer, P. 2002.
\newblock {Finite}-{Time} {Analysis} of the {Multiarmed} {Bandit} {Problem}.
\newblock \emph{Machine learning}, 47(2-3): 235--256.

\bibitem[{Badia et~al.(2020)Badia, Piot, Kapturowski, Sprechmann, Vitvitskyi,
  Guo, and Blundell}]{badia2020agent57}
Badia, A.~P.; Piot, B.; Kapturowski, S.; Sprechmann, P.; Vitvitskyi, A.; Guo,
  Z.~D.; and Blundell, C. 2020.
\newblock {Agent57}: {Outperforming} the {Atari} {Human} {Benchmark}.
\newblock In \emph{International conference on machine learning}, 507--517.
  PMLR.

\bibitem[{Bai, Wu, and Chen(2013)}]{bai2013bayesian}
Bai, A.; Wu, F.; and Chen, X. 2013.
\newblock {Bayesian} {Mixture} {Modelling} and {Inference} based {Thompson}
  {Sampling} in {Monte}-{Carlo} {Tree} {Search}.
\newblock In \emph{Advances in Neural Information Processing Systems},
  1646--1654.

\bibitem[{Bai et~al.(2014)Bai, Wu, Zhang, and Chen}]{bai2014thompson}
Bai, A.; Wu, F.; Zhang, Z.; and Chen, X. 2014.
\newblock {Thompson} {Sampling} based {Monte}-{Carlo} {Planning} in {POMDPs}.
\newblock In \emph{Proceedings of the Twenty-Fourth International Conferenc on
  International Conference on Automated Planning and Scheduling}, 29--37. AAAI
  Press.

\bibitem[{Chapelle and Li(2011)}]{chapelle2011empirical}
Chapelle, O.; and Li, L. 2011.
\newblock {An} {Empirical} {Evaluation} of {Thompson} {Sampling}.
\newblock In \emph{Advances in neural information processing systems},
  2249--2257.

\bibitem[{Chen and Bai(2018)}]{Chen2018ARL}
Chen, B.; and Bai, R. 2018.
\newblock {A} {Reinforcement} {Learning} {Based} {Variable} {Neighborhood}
  {Search} {Algorithm} for {Open} {Periodic} {Vehicle} {Routing} {Problem} with
  {Time} {Windows}.

\bibitem[{Chen et~al.(2016)Chen, Cowling, Polack, and Mourdjis}]{Chen2016AMB}
Chen, Y.; Cowling, P.~I.; Polack, F. A.~C.; and Mourdjis, P. 2016.
\newblock {A} {Multi}-{Arm} {Bandit} {Neighbourhood} {Search} for {Routing} and
  {Scheduling} {Problems}.

\bibitem[{Chmiela et~al.(2023)Chmiela, Gleixner, Lichocki, and
  Pokutta}]{chmiela2023online}
Chmiela, A.; Gleixner, A.; Lichocki, P.; and Pokutta, S. 2023.
\newblock {Online} {Learning} for {Scheduling} {MIP} {Heuristics}.
\newblock In \emph{International Conference on Integration of Constraint
  Programming, Artificial Intelligence, and Operations Research}, 114--123.
  Springer.

\bibitem[{Cohen et~al.(2018)Cohen, Greco, Ma, Hern{\'a}ndez, Felner, Kumar, and
  Koenig}]{cohen2018anytime}
Cohen, L.; Greco, M.; Ma, H.; Hern{\'a}ndez, C.; Felner, A.; Kumar, T.~S.; and
  Koenig, S. 2018.
\newblock {Anytime} {Focal} {Search} with {Applications}.
\newblock In \emph{IJCAI}, 1434--1441.

\bibitem[{Garivier and Moulines(2008)}]{garivier2008upper}
Garivier, A.; and Moulines, E. 2008.
\newblock {On} {Upper}-{Confidence} {Bound} {Policies} for {Non}-{Stationary}
  {Bandit} {Problems}.
\newblock \emph{arXiv preprint arXiv:0805.3415}.

\bibitem[{Goldberg(1988)}]{Goldberg1988GeneticAI}
Goldberg, D.~E. 1988.
\newblock {Genetic} {Algorithms} in {Search} {Optimization} and {Machine}
  {Learning}.

\bibitem[{Hendel(2022)}]{hendel2022adaptive}
Hendel, G. 2022.
\newblock {Adaptive} {Large} {Neighborhood} {Search} for {Mixed} {Integer}
  {Programming}.
\newblock \emph{Mathematical Programming Computation}, 1--37.

\bibitem[{Huang, Dilkina, and Koenig(2021)}]{huang2021learning}
Huang, T.; Dilkina, B.; and Koenig, S. 2021.
\newblock {Learning} {Node}-{Selection} {Strategies} in {Bounded} {Suboptimal}
  {Conflict}-{Based} {Search} for {Multi}-{Agent} {Path} {Finding}.
\newblock In \emph{International Joint Conference on Autonomous Agents and
  Multiagent Systems (AAMAS)}.

\bibitem[{Huang et~al.(2022)Huang, Li, Koenig, and Dilkina}]{HuangAAAI22}
Huang, T.; Li, J.; Koenig, S.; and Dilkina, B. 2022.
\newblock {Anytime} {Multi}-{Agent} {Path} {Finding} via {Machine}
  {Learning}-{Guided} {Large} {Neighborhood} {Search}.
\newblock In \emph{Proceedings of the 36th AAAI Conference on Artificial
  Intelligence (AAAI)}, 9368--9376.

\bibitem[{Kaduri, Boyarski, and Stern(2020)}]{kaduri2020algorithm}
Kaduri, O.; Boyarski, E.; and Stern, R. 2020.
\newblock {Algorithm} {Selection} for {Optimal} {Multi}-{Agent} {Pathfinding}.
\newblock In \emph{Proceedings of the International Conference on Automated
  Planning and Scheduling}, volume~30, 161--165.

\bibitem[{Kaufmann, Korda, and Munos(2012)}]{kaufmann2012thompson}
Kaufmann, E.; Korda, N.; and Munos, R. 2012.
\newblock {Thompson} {Sampling}: {An} {Asymptotically} {Optimal}
  {Finite}-{Time} {Analysis}.
\newblock In \emph{International Conference on Algorithmic Learning Theory},
  199--213. Springer.

\bibitem[{Kocsis and Szepesv{\'a}ri(2006)}]{kocsis2006bandit}
Kocsis, L.; and Szepesv{\'a}ri, C. 2006.
\newblock {Bandit} based {Monte}-{Carlo} {Planning}.
\newblock In \emph{ECML}, volume~6, 282--293. Springer.

\bibitem[{Lam et~al.(2023)Lam, Harabor, Stuckey, and Li}]{LamICAPS23}
Lam, E.; Harabor, D.; Stuckey, P.~J.; and Li, J. 2023.
\newblock {Exact} {Anytime} {Multi}-{Agent} {Path} {Finding} {Using}
  {Branch}-and-{Cut}-and-{Price} and {Large} {Neighborhood} {Search}.
\newblock In \emph{Proceedings of the International Conference on Automated
  Planning and Scheduling (ICAPS)}.

\bibitem[{Li et~al.(2021)Li, Chen, Harabor, Stuckey, and
  Koenig}]{li2021anytime}
Li, J.; Chen, Z.; Harabor, D.; Stuckey, P.~J.; and Koenig, S. 2021.
\newblock {Anytime} {Multi}-{Agent} {Path} {Finding} via {Large} {Neighborhood}
  {Search}.
\newblock In \emph{Proceedings of the International Joint Conference on
  Artificial Intelligence (IJCAI)}, 4127--4135.

\bibitem[{Li et~al.(2022)Li, Chen, Harabor, Stuckey, and Koenig}]{li2022lns2}
Li, J.; Chen, Z.; Harabor, D.; Stuckey, P.~J.; and Koenig, S. 2022.
\newblock {MAPF}-{LNS2}: {Fast} {Repairing} for {Multi}-{Agent} {Path}
  {Finding} via {Large} {Neighborhood} {Search}.
\newblock \emph{Proceedings of the AAAI Conference on Artificial Intelligence},
  36(9): 10256--10265.

\bibitem[{Mara et~al.(2022)Mara, Norcahyo, Jodiawan, Lusiantoro, and
  Rifai}]{mara2022survey}
Mara, S. T.~W.; Norcahyo, R.; Jodiawan, P.; Lusiantoro, L.; and Rifai, A.~P.
  2022.
\newblock {A} {Survey} of {Adaptive} {Large} {Neighborhood} {Search}
  {Algorithms} and {Applications}.
\newblock \emph{Computers \& Operations Research}, 146: 105903.

\bibitem[{Phan et~al.(2019{\natexlab{a}})Phan, Belzner, Kiermeier, Friedrich,
  Schmid, and Linnhoff-Popien}]{phan2019memory}
Phan, T.; Belzner, L.; Kiermeier, M.; Friedrich, M.; Schmid, K.; and
  Linnhoff-Popien, C. 2019{\natexlab{a}}.
\newblock {Memory} {Bounded} {Open}-{Loop} {Planning} in {Large} {POMDPs} using
  {Thompson} {Sampling}.
\newblock \emph{Proceedings of the AAAI Conference on Artificial Intelligence},
  33(01): 7941--7948.

\bibitem[{Phan et~al.(2019{\natexlab{b}})Phan, Gabor, Müller, Roch, and
  Linnhoff-Popien}]{phan2019adaptive}
Phan, T.; Gabor, T.; Müller, R.; Roch, C.; and Linnhoff-Popien, C.
  2019{\natexlab{b}}.
\newblock {Adaptive} {Thompson} {Sampling} {Stacks} for {Memory} {Bounded}
  {Open}-{Loop} {Planning}.
\newblock In \emph{Proceedings of the 28th International Joint Conference on
  Artificial Intelligence, {IJCAI-19}}, 5607--5613. International Joint
  Conferences on Artificial Intelligence Organization.

\bibitem[{Ratner and Warmuth(1986)}]{ratner1986finding}
Ratner, D.; and Warmuth, M. 1986.
\newblock {Finding} a {Shortest} {Solution} for the {N}x{N} {Extension} of the
  15-{Puzzle} is {Intractable}.
\newblock In \emph{Proceedings of the Fifth AAAI National Conference on
  Artificial Intelligence}, AAAI'86, 168–172. AAAI Press.

\bibitem[{Ropke and Pisinger(2006)}]{ropke2006adaptive}
Ropke, S.; and Pisinger, D. 2006.
\newblock {An} {Adaptive} {Large} {Neighborhood} {Search} {Heuristic} for the
  {Pickup} and {Delivery} {Problem} with {Time} {Windows}.
\newblock \emph{Transportation science}, 40(4): 455--472.

\bibitem[{Rothberg(2007)}]{rothberg2007evolutionary}
Rothberg, E. 2007.
\newblock {An} {Evolutionary} {Algorithm} for {Polishing} {Mixed} {Integer}
  {Programming} {Solutions}.
\newblock \emph{INFORMS Journal on Computing}, 19(4): 534--541.

\bibitem[{Sartoretti et~al.(2019)Sartoretti, Kerr, Shi, Wagner, Kumar, Koenig,
  and Choset}]{sartoretti2019primal}
Sartoretti, G.; Kerr, J.; Shi, Y.; Wagner, G.; Kumar, T.~S.; Koenig, S.; and
  Choset, H. 2019.
\newblock {PRIMAL}: {Pathfinding} via {Reinforcement} and {Imitation}
  {Multi}-{Agent} {Learning}.
\newblock \emph{IEEE Robotics and Automation Letters}, 4(3): 2378--2385.

\bibitem[{Schaul et~al.(2019)Schaul, Borsa, Ding, Szepesvari, Ostrovski,
  Dabney, and Osindero}]{schaul2019adapting}
Schaul, T.; Borsa, D.; Ding, D.; Szepesvari, D.; Ostrovski, G.; Dabney, W.; and
  Osindero, S. 2019.
\newblock {Adapting} {Behaviour} for {Learning} {Progress}.
\newblock \emph{arXiv preprint arXiv:1912.06910}.

\bibitem[{Sharon et~al.(2012)Sharon, Stern, Felner, and
  Sturtevant}]{sharon2012conflict}
Sharon, G.; Stern, R.; Felner, A.; and Sturtevant, N. 2012.
\newblock {Conflict}-{Based} {Search} {For} {Optimal} {Multi}-{Agent} {Path}
  {Finding}.
\newblock \emph{Proceedings of the AAAI Conference on Artificial Intelligence},
  26(1): 563--569.

\bibitem[{Silver(2005)}]{silver2005cooperative}
Silver, D. 2005.
\newblock {Cooperative} {Pathfinding}.
\newblock \emph{Proceedings of the AAAI Conference on Artificial Intelligence
  and Interactive Digital Entertainment}, 1(1): 117--122.

\bibitem[{Silver and Veness(2010)}]{silver2010monte}
Silver, D.; and Veness, J. 2010.
\newblock {Monte}-{Carlo} {Planning} in {Large} {POMDPs}.
\newblock In \emph{Advances in neural information processing systems},
  2164--2172.

\bibitem[{Stern et~al.(2019)Stern, Sturtevant, Felner, Koenig, Ma, Walker, Li,
  Atzmon, Cohen, Kumar et~al.}]{stern2019multi}
Stern, R.; Sturtevant, N.; Felner, A.; Koenig, S.; Ma, H.; Walker, T.; Li, J.;
  Atzmon, D.; Cohen, L.; Kumar, T.; et~al. 2019.
\newblock {Multi}-{Agent} {Pathfinding}: {Definitions}, {Variants}, and
  {Benchmarks}.
\newblock In \emph{Proceedings of the International Symposium on Combinatorial
  Search}, volume~10, 151--158.

\bibitem[{{\'S}wiechowski et~al.(2023){\'S}wiechowski, Godlewski, Sawicki, and
  Ma{\'n}dziuk}]{swiechowski2023monte}
{\'S}wiechowski, M.; Godlewski, K.; Sawicki, B.; and Ma{\'n}dziuk, J. 2023.
\newblock {Monte} {Carlo} {Tree} {Search}: {A} {Review} of {Recent}
  {Modifications} and {Applications}.
\newblock \emph{Artificial Intelligence Review}, 56(3): 2497--2562.

\bibitem[{Thompson(1933)}]{thompson1933likelihood}
Thompson, W.~R. 1933.
\newblock {On} the {Likelihood} that {One} {Unknown} {Probability} exceeds
  {Another} in {View} of the {Evidence} of {Two} {Samples}.
\newblock \emph{Biometrika}, 25(3/4): 285--294.

\bibitem[{Yu and LaValle(2013)}]{yu2013structure}
Yu, J.; and LaValle, S. 2013.
\newblock {Structure} and {Intractability} of {Optimal} {Multi}-{Robot} {Path}
  {Planning} on {Graphs}.
\newblock \emph{Proceedings of the AAAI Conference on Artificial Intelligence},
  27(1): 1443--1449.

\end{thebibliography}

\end{document}